\documentclass{article}
\usepackage[preprint]{neurips_2025}

\usepackage{graphicx} 
\usepackage[utf8]{inputenc} 
\usepackage[T1]{fontenc}    
\usepackage[colorlinks=true, linkcolor=blue, citecolor=blue, urlcolor=blue]{hyperref}
\usepackage{url}            
\usepackage{booktabs}       
\usepackage{amsfonts}       
\usepackage{nicefrac}       
\usepackage{microtype}      
\usepackage{xcolor}         
\usepackage{caption}

\graphicspath{{./figs/}}
\newcommand{\tool}{\textit{SAE Semantic Explorer}}
\newcommand{\para}[1]{\noindent{{\textbf{#1}}}}
\title{Visual Exploration of Feature Relationships in Sparse Autoencoders with Curated Concepts}

\author{Xinyuan Yan, Shusen Liu, Kowshik Thopalli, Bei Wang} 

\author{
  Xinyuan Yan\\
  University of Utah\\
  \texttt{xinyuan.yan@utah.edu} \\
  \And
  Shusen Liu \\
  Lawrence Livermore National Laboratory \\
  \texttt{liu42@llnl.gov} \\
  \AND
  Kowshik Thopalli \\
  Lawrence Livermore National Laboratory \\
  \texttt{thopalli1@llnl.gov} \\
  \And
  Bei Wang \\
  University of Utah \\
  \texttt{beiwang@sci.utah.edu} \\
}

\begin{document}

\maketitle

\begin{abstract}
Sparse autoencoders (SAEs) have emerged as a powerful tool for uncovering interpretable features in large language models by learning sparse directions in their activation spaces. However, the sheer number of extracted directions renders comprehensive exploration intractable. Conventional embedding methods such as UMAP can reveal global organization but often introduce high-dimensional compression artifacts, overplotting, and misleading neighborhood distortions.
In this work, we introduce a focused exploration framework that prioritizes curated concepts and their associated SAE features over exhaustive visualization of all features. We present an interactive visualization system that integrates topology-based visual encodings with dimensionality reduction to preserve both local and global relationships among selected features. This hybrid approach enables targeted, interpretable analysis of SAE behavior, supporting deeper insight into how concepts are represented across layers of large language models.

\end{abstract}

\begin{figure}
\centering
\vspace{-10mm}
\includegraphics[width=\linewidth]{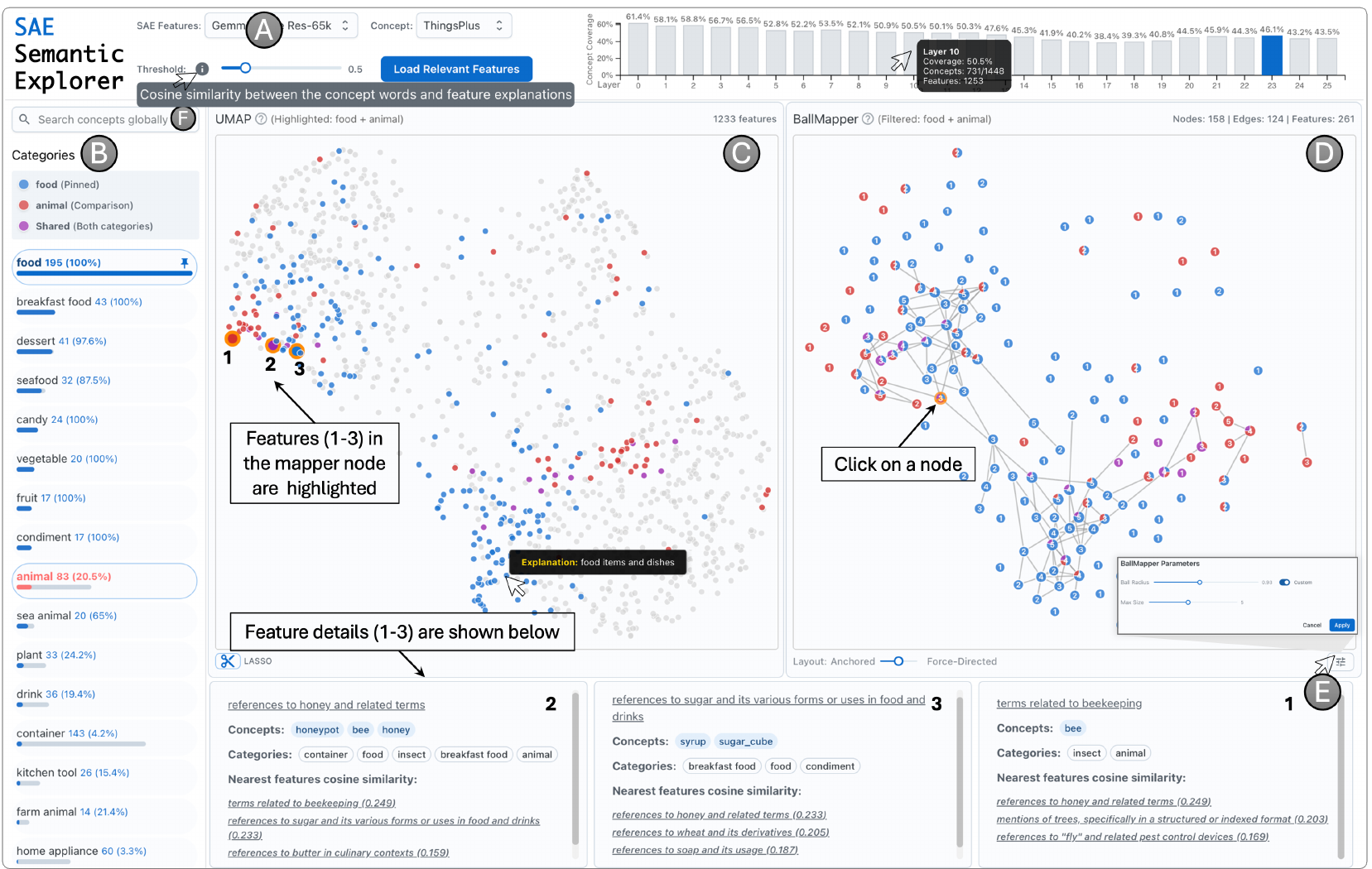}
\vspace{-2mm}
\captionof{figure}{
{\tool} interface.
A. \textbf{Data view}. Left: SAE features, a concept set (words with assigned categories), and a cosine-similarity threshold for retrieving relevant features. Right: bar chart showing the number of discovered concepts per layer.
B. \textbf{Category view}. For the selected layer (23), each row displays a category’s feature count and its overlap with the pinned category \emph{food}, facilitating comparison with \emph{animal}.
C. \textbf{UMAP view}. Retrieved features from the selected layer, with \emph{food} and \emph{animal} categories highlighted.
D. \textbf{Ball Mapper view}. Topological graph showing the structural relationships among \emph{food} and \emph{animal}  features.
E. \textbf{Feature view}. Interactive panel displaying details of selected features via click or lasso selection.
F. \textbf{Concept query}. Search interface for locating specific concepts.}
\label{fig:interface}
\vspace{-4mm}
\end{figure}

\section{Introduction}
\label{sec:introduction}

Sparse autoencoders (SAEs) have emerged as a powerful technique for extracting interpretable features from large language models (LLMs), decomposing superposed neural representations into disentangled components~\cite{sharkey2022taking, elhage2022toy, bricken2023towards, huben2023sparse}. Recent work has demonstrated remarkable scalability, extracting millions of interpretable features from state-of-the-art models~\cite{templeton2024scaling, gao2024scaling, balcells2024evolution}. Yet this success introduces a paradox: the sheer number of learned sparse directions, often hundreds of thousands to millions, renders comprehensive exploration both computationally and cognitively intractable. Moreover, recent studies~\cite{engels2024not, rajamanoharan2024improving} reveal that many SAE features are \emph{polysemantic} (encoding multiple, unrelated concepts) or low-quality, raising fundamental questions about whether visualizing all features simultaneously is even meaningful.

In this work, we advocate a \emph{focused exploration} paradigm that emphasizes carefully curated concept sets and their corresponding SAE features, rather than attempting to visualize all learned features at once~\cite{bloom2024saetrainingcodebase, neuronpedia, joseph2025prisma}. By concentrating on well-defined concepts, researchers can avoid the confusion introduced by the vast number of noisy or potentially low-quality features and instead conduct targeted, hypothesis-driven investigations into concept representation and feature relationships.

We pursue three interconnected analytical objectives that are central to understanding the conceptual organization of SAE features.
First, we examine how cosine similarity among features corresponds to semantic similarity, providing a quantitative validation of whether SAEs capture meaningful semantic structure~\cite{marks2024sparse}.
Second, we evaluate how human-curated concept hierarchies are reflected in the SAE feature space, assessing whether models exhibit organizational patterns consistent with human understanding~\cite{ inaba2025bilingual, boggust2025abstraction}.
Third, we analyze how both local and global structures evolve across model layers, as feature representations undergo systematic transformations with network depth~\cite{balcells2024evolution, lawson2024residual}.

Existing SAE visualizations commonly employ dimensionality reduction methods such as UMAP~\cite{mcinnes2018umap}, which compress thousands of dimensions into two. However, such extreme compression often introduces severe structural distortions that misrepresent neighborhood relationships~\cite{liu2014distortion}, which are critical for interpretability and understanding semantic similarities between features. For example, similar features may appear artificially separated, while dissimilar ones may appear spuriously clustered, leading to misleading conclusions about feature organization.

To address these limitations, we introduce a topology-based visual encoding inspired by topological data analysis. Building on the ball mapper algorithm~\cite{dlotko2019ball}, our method constructs a network representation that preserves both local and global structural properties. Unlike projection-based embeddings that force features into 2D space, our topological representation maintains discrete feature clusters and explicitly encodes their interconnections, providing a more faithful depiction of similarity relationships among SAE features.

Finally, we present {\tool}, a visual analytics tool that integrates this topology-based encoding with dimensionality reduction techniques. Through multiple coordinated views, our tool enables interactive exploration of local, global, and cross-layer feature relationships, supporting flexible and hypothesis-driven analysis of SAEs. The code, datasets, and video demonstration are publicly available at  \url{https://github.com/tdavislab/SAEExploration.git}.

\section{Related Works}
\label{sec:related-works}

\paragraph{Foundations, evaluation, and analysis of SAEs.}
Sparse autoencoders (SAEs) have emerged as an unsupervised approach for decomposing latent representations into approximately monosemantic and interpretable features~\cite{makhzani2013k, zhang2016survey}.
Early studies demonstrate that SAEs can recover interpretable directions in both toy transformers and LLMs~\cite{sharkey2022taking, elhage2022toy, huben2023sparse}.
Subsequent research has advanced and scaled SAEs through architectural modifications~\cite{rajamanoharan2024improving}, alternative activation functions~\cite{gao2024scaling, rajamanoharan2024jumping}, and enhanced loss formulations~\cite{marks2024enhancing, bussmann2025learning}.
To interpret individual features, automated methods employ LLMs to summarize the tokens that most strongly activate each feature~\cite{rajamanoharan2024jumping, paulo2025automatically}, while SAEBench~\cite{karvonen2025saebench} provides a standardized benchmark for comprehensive quantitative evaluation.
Complementary studies further analyze how SAE features evolve across network layers~\cite{balcells2024evolution, inaba2025bilingual} and across model scales~\cite{leask2025sparse}, revealing characteristic trends in representation structure.
Despite these interpretability gains, recent findings highlight instability in the learned features~\cite{paulo2025sparse} and show that SAEs may underperform simpler baselines in representation steering tasks~\cite{wu2025axbench}, raising questions about their practical reliability.

\paragraph{Visualizing SAE features.}
An emerging ecosystem of interactive visualization platforms supports the exploration and manipulation of SAE-learned features, including SAELens~\cite{bloom2024saetrainingcodebase}, Neuronpedia~\cite{neuronpedia}, and the interactive analyses presented in Anthropic’s interpretability reports~\cite{bricken2023monosemanticity}.
These systems demonstrate both the utility and the usability challenges of exploring thousands to millions of discovered directions.
Such challenges are compounded by the use of dimensionality-reduction methods like UMAP~\cite{mcinnes2018umap}, where distance distortion and overplotting can obscure meaningful structure~\cite{liu2014distortion}.
In this work, we instead enable a more \emph{focused} exploration of selected, conceptually meaningful features through curated concepts and a topology-driven analysis that faithfully captures structural relationships.

\paragraph{Mapper graph for model interpretability.}
The mapper graph~\cite{SinghMemoliCarlsson2007} is a topological data analysis technique that summarizes the structure of a point cloud as a graph, where each node represents a cluster of points, and edges connect nodes with overlapping members.
Mapper graphs have recently been leveraged to interpret model embeddings by revealing the topology of latent spaces. For instance, they have been used to capture the structural organization of hidden representations in image classifiers and large language models (LLMs)~\cite{yan2025explainable, RathoreZhouSrikumar2023, RathoreChalapathiPalande2021}, to analyze how representations evolve across layers and during fine-tuning~\cite{yan2025explainable, RathoreZhouSrikumar2023}, and to characterize the effects of adversarial perturbations~\cite{ZhouZhouDing2023}.
In this work, we employ \emph{ball mapper}~\cite{dlotko2019ball}, a variant of mapper with fewer parameters, to complement dimensionality reduction methods, providing a view that better preserves both local neighborhood relations and global structural patterns.

\section{{\tool}: Interactive Concept Exploration}
\label{sec:tool}

We develop a visual analytics tool, {\tool}, that enables human-in-the-loop exploration of SAE features (Figure~\ref{fig:interface}). 

\para{Data description and preprocessing.}
The input data comprise SAE features (high-dimensional vectors) across layers, LLM-generated textual explanations for each feature~\cite{caden2024open}, and a concept dataset containing concept words and their associated categories (e.g., the category \emph{animal} includes \emph{dog}, \emph{cat}, etc.; a concept may belong to multiple categories).
For each layer, relevant features for each concept are retrieved via the Neuronpedia API~\cite{neuronpedia}, which computes cosine similarity between concept and feature explanations in a sentence embedding space.
By default, features with similarity scores above $0.5$ are visualized, though users can adjust this threshold interactively.

\begin{figure}
\centering
\includegraphics[width=0.9\linewidth]{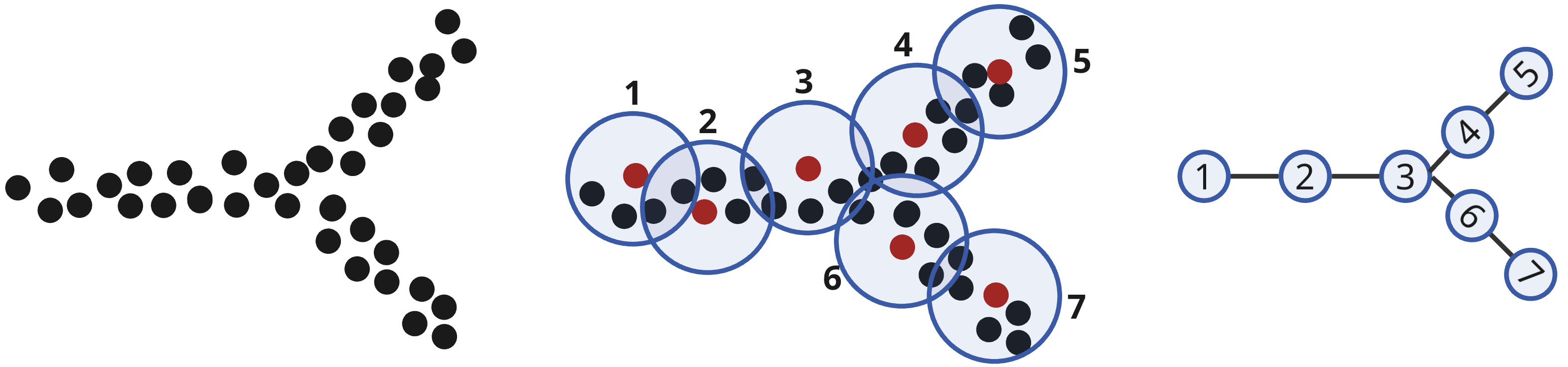}
\vspace{-2mm}
\captionof{figure}{
Ball mapper construction example.
Left: Original point cloud.
Middle: For a given radius $\epsilon$, a subset of points (red) is selected as ball centers such that the resulting balls (1–7, blue) cover the entire dataset.
Right: The resulting ball mapper graph represents each ball as a node, with an edge between two nodes if their corresponding balls share data points.}
\label{fig:ballmapper}
\vspace{-4mm}
\end{figure}

\para{Embedding view based on ball mapper.}
To better preserve local structure and mitigate distortions introduced by traditional dimensionality reduction, we introduce an embedding view grounded in the ball mapper framework.
Ball mapper~\cite{dlotko2019ball} encodes the topology of a high-dimensional point cloud as a graph, where nodes represent local neighborhoods (balls) of points and edges denote overlaps between them.
As illustrated in Figure~\ref{fig:ballmapper}, given a user-defined radius~$\epsilon$, a subset of points is selected (e.g., via a greedy procedure) such that balls centered at these points cover the entire dataset. Each ball defines a node, and two nodes are connected if their corresponding balls share points.
By default, $\epsilon$ is estimated by computing all pairwise cosine distances and selecting the elbow point of the resulting distribution, following common practice~\cite{zhou2021mapper}. 
To reduce visual clutter, we employ an adaptive variant: for a specified maximum node size, the ball radius is iteratively reduced by a factor~$\eta$ (default~0.9) until the constraint is satisfied. In our system, the maximum node size defaults to~5, though users can adjust this and other parameters interactively.

\para{Interface design.}
The interface integrates multiple coordinated views. As shown in Figure~\ref{fig:interface}, View A allows users to select SAE feature data, concept data, and a similarity threshold to retrieve concept-relevant features per layer. The resulting distribution of discovered concepts is displayed as a bar chart; clicking a bar loads the corresponding layer data in Views B–D.

For a selected layer, View B lists category information, including the number of relevant features associated with each category. View C presents a UMAP projection of features, supporting zoom, pan, hover, lasso, and selection, with the three nearest features highlighted on click. View D shows the ball mapper view, where users can drag, adjust parameters, and select nodes or edges. The ball mapper graph supports a smooth transition between two layouts: an \emph{anchored} layout, in which nodes are positioned by the average location of their UMAP points for alignment, and a \emph{force-directed} layout for a clearer, aesthetically pleasing visualization.

View E displays detailed information about selected features, including textual explanations, related concepts and categories, and nearby features within the same layer. Each explanation links directly to the Neuronpedia feature detail page~\cite{neuronpedia}. View F provides a search interface for retrieving features associated with specific concepts.

To aid navigation, tooltips are available throughout the interface. Selecting a category in View B highlights its corresponding features in both the UMAP and ball mapper views. Users can pin a category to explore relationships among categories, after which others are sorted by shared features. When a comparison category is selected, features are consistently color-coded across UMAP and ball mapper views to reveal overlaps and distinctions.

\section{Exploratory Analysis Results}
\label{sec:results}
In this section, we present insights obtained through the proposed visualization framework, highlighting how it facilitates understanding of the relationships between learned SAE features and human-interpretable concepts.

\para{Concept sets.}
We analyze two hierarchical concept sets: (i) \texttt{THINGSplus}~\cite{stoinski2024thingsplus}, a human-curated dataset containing $1{,}448$ concepts organized into $53$ categories (e.g., \emph{mammals}, \emph{food}), where each category includes a list of relevant concepts (e.g., \emph{panda}, \emph{cat}, \emph{breadsticks}); and (ii) \texttt{Subjects}, a field-of-study-oriented set curated via an LLM, containing $1{,}683$ concepts spanning eight disciplines (e.g., \emph{mathematics}, \emph{physics}) and their subtopics (e.g., \emph{algebra}, \emph{geometry}, \emph{calculus} for mathematics).
While most SAE visualization frameworks~\cite{gao2024scaling} focus on providing overviews of learned features, they often lack mechanisms for connecting those features to practitioner-relevant concepts. Our framework addresses this gap by allowing users to load concept sets and retrieve features aligned with specific concepts of interest, enabling targeted exploration.
To demonstrate this capability, we analyze 65k open-source SAE features from each layer of the \texttt{gemma-2-2b} model~\cite{team2024gemma}, as provided by Gemma Scope~\cite{lieberum2024gemma}. Neuronpedia~\cite{neuronpedia} supplies corresponding textual explanations for each feature, generated via the Auto-Interp framework~\cite{paulo2025automatically}, across 26 residual-stream layers.

\para{Evolution of concept relationships across layers.}
Using our visualization framework, we observe several patterns in how concept relationships evolve across layers.
For the \texttt{THINGSplus} concept set, early layers produce UMAP representations that are densely clustered; these gradually separate into two major clusters in the middle layers and later recombine into more integrated structures in deeper layers. This trajectory suggests that the model first learns broad, general representations, refines them into specialized concepts, and subsequently integrates them into cohesive, high-level abstractions—a pattern consistent with prior findings on the evolution of factual knowledge in LLMs~\cite{meng2022locating}.
The ball mapper view, however, reveals a more nuanced picture: multiple smaller clusters emerge in the middle layers, indicating a diversity of feature specialization not fully captured by UMAP embedding.

\begin{figure}[!ht]
\vspace{-2mm}
\centering
\includegraphics[width=\linewidth]{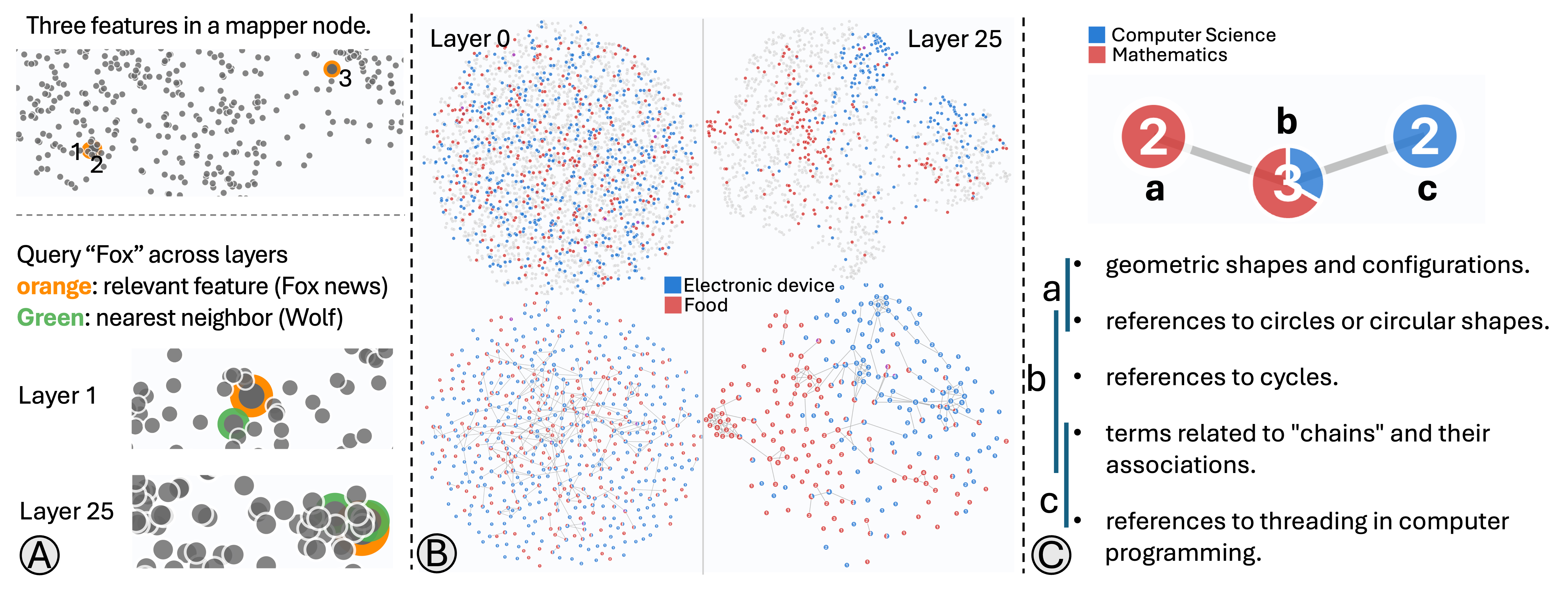}
\vspace{-5mm}
\caption{
\textbf{A.} Top: a mapper node containing features 1–3, all related to the concept \emph{music album}; bottom: querying \emph{fox} across layers consistently yields \emph{Fox News}, with \emph{wolf} as its nearest neighbor.
\textbf{B.} UMAP and ball mapper views for layers 0 and 25, highlighting features associated with \emph{food} and \emph{electronic devices}.
\textbf{C.} A ball mapper path of subject concepts illustrating a transition from \emph{Mathematics} to \emph{Computer Science}.
}
\vspace{-4mm}
\label{fig:results}
\end{figure}

For the \texttt{Subjects} concept set, this layer-wise differentiation is less pronounced: the UMAP embeddings remain relatively clustered, likely due to the abstract nature of academic domains, which exhibit weaker separability in feature space.
Additionally, ball mapper neighborhoods highlight persistent local semantic relationships across layers, consistent with prior observations~\cite{balagansky2024mechanistic}. For instance, in Figure~\ref{fig:results}(A, bottom), querying \emph{fox} consistently retrieves features related to \emph{Fox News}, with \emph{wolf} appearing as its nearest neighbor within the same mapper node.

\para{Complementary local and global analysis.}
Our tool supports complementary local and global analyses of the feature space: ball mapper view exposes topological structure and reveals distortions introduced by dimensionality reduction that may be obscured in the UMAP view.
This complementarity is illustrated in Figure~\ref{fig:results}(A, top), where features~1–3 within a ball mapper node all correspond to the concept \emph{music album}. While features~1 and~2 are true nearest neighbors in the original space, their proximity is distorted in the UMAP view. As discussed earlier, our visualization also shows that local semantic proximity often persists across layers.

At the global scale, Figure~\ref{fig:results}(B) highlights category features associated with \emph{electronic devices} and \emph{food}. In early layers, these features appear mixed and spatially dispersed in both views, whereas deeper layers exhibit clearer intra-category grouping and stronger inter-category separation. Overlaps between categories also emerge; for instance, Figure~\ref{fig:interface}(B) shows food-related features intersecting with \emph{breakfast} and \emph{dessert}, while Figure~\ref{fig:interface}(D) reveals a ball mapper node containing \emph{sugar}, \emph{honey}, and \emph{bee}, effectively marking a semantic boundary between \emph{food} and \emph{animal}.

Our framework further enables exploration of transitions between related concepts. In Figure~\ref{fig:results}(C), a path in the ball mapper graph (i.e., a sequence of overlapping local neighborhoods) traces a progression from \emph{Computer Science} to \emph{Mathematics}, illustrating how the model encodes relationships across fields of study. This observation aligns with prior work showing that mapper graphs can capture the evolution of linguistic phenomena along trajectories in embedding space~\cite{RathoreZhouSrikumar2023}.

Although we present only a few representative cases, these examples demonstrate that our framework provides a scalable and interpretable means of exploring how user-defined concepts relate to the structured knowledge encoded in SAEs across layers. We anticipate that such targeted exploration will facilitate deeper insight into model organization and support downstream applications such as feature-level intervention~\cite{chalnev2024improving}, model editing~\cite{wang2024knowledge}, and unlearning~\cite{muhamed2025saes}.


\para{Limitations and discussion.}
Despite our exploration of curated concepts, a key limitation remains the inherent challenge of \emph{auto-interpretability} in SAE features, that is, the reliability of identified features depends heavily on the quality of their automatically generated explanations.
We plan to enhance the UMAP visualization by integrating scalable and annotated visualization toolkits~\cite{wang-etal-2023-wizmap, ren2025embedding}, and to leverage LLMs to automatically generate richer, context-aware explanations for mapper nodes~\cite{yan2025explainable}.

\begin{ack}
This work was partially supported by the U.S. National Science Foundation (NSF) under grants DMS-2134223 and IIS-2205418. It was performed under the auspices of the U.S. Department of Energy (DOE) by Lawrence Livermore National Laboratory under Contract DE-AC52-07NA27344, with additional support from the LLNL LDRD program (23-ERD-029, 25-SI-001) and the DOE Early Career Research Program (ECRP, SCW1885). This work has been reviewed and released under LLNL-CONF-2010620.
\end{ack}

\bibliographystyle{unsrt}
\bibliography{refs}

\end{document}